\documentclass[conference]{IEEEtran}

\usepackage[pdftex]{graphicx}
\usepackage{amsfonts}
\usepackage[cmex10]{amsmath}
\usepackage{mathabx}
\usepackage{algorithmic}
\usepackage{array}
\usepackage{mdwmath}
\usepackage{mdwtab}
\usepackage{eqparbox}
\usepackage{url}
\usepackage{subcaption}
\usepackage{float}
\usepackage{booktabs,chemformula}
\usepackage{hyperref}
\usepackage{caption}
\usepackage{cleveref}
\usepackage{fancyhdr}
\usepackage[super]{nth}
\usepackage[pscoord]{eso-pic}
\usepackage{amssymb}
\usepackage{amsmath}

\makeatletter
\newcommand\AtPageUpperMyright[1]{\AtPageUpperLeft{
     \put(\LenToUnit{0.5\paperwidth},\LenToUnit{-1cm}){
         \parbox{0.5\textwidth}{\raggedleft\fontsize{9}{11}\selectfont #1}}
     }}
\newcommand{\confheader}[1]{
    \AddToShipoutPictureBG*{
    \AtPageUpperMyright{#1}
    }
}
\def\endthebibliography{%
	\def\@noitemerr{\@latex@warning{Empty `thebibliography' environment}}%
	\endlist
}
\makeatother

\begin{document}
	\title{Jamdani Motif Generation using Conditional GAN}
	% \author{\authorblockN{Leave Author List blank for your IMS2013 Summary (initial) submission.\\ IMS2013 will be rigorously enforcing the new double-blind reviewing requirements.}
	% \authorblockA{\authorrefmark{1}Leave Affiliation List blank for your Summary (initial) submission}}
	
	\author{\authorblockN{
	MD Tanvir Rouf Shawon\authorrefmark{1} \textsuperscript{\textsection}, 
	Raihan Tanvir\authorrefmark{2}
	\textsuperscript{\textsection}, 
	Humaira Ferdous Shifa\authorrefmark{3}, 
	Susmoy Kar\authorrefmark{5}, 
	Mohammad Imrul Jubair\authorrefmark{6}  }
 	\authorblockA{Department of Computer Science and Engineering}
 	\authorblockA{Ahsanullah University of Science and Technology Dhaka, Bangladesh}
	\authorblockA{\{\authorrefmark{1}shawontanvir95, \authorrefmark{2}raihantanvir.96, \authorrefmark{3}yusufhumairaechad, \authorrefmark{5}susmoy.sun16\}@gmail.com, \authorrefmark{6}mohammadimrul.jubair@ucalgary.ca}}
	    
	\maketitle
	\begingroup\renewcommand\thefootnote{\textsection}
        \footnotetext{MD Tanvir Rouf Shawon and Raihan Tanvir have equal contributions.}
    \endgroup
    
	\begin{abstract}
		Jamdani is the strikingly patterned textile heritage of Bangladesh. The exclusive geometric motifs woven on the fabric are the most attractive part of this craftsmanship having a remarkable influence on textile and fine art. In this paper, we have developed a technique based on the Generative Adversarial Network that can learn to generate entirely new Jamdani patterns from a collection of Jamdani motifs that we assembled, the newly formed motifs can mimic the appearance of the original designs. Users can input the skeleton of a desired pattern in terms of rough strokes and our system finalizes the input by generating the complete motif which follows the geometric structure of real Jamdani ones. To serve this purpose, we collected and preprocessed a dataset containing a large number of Jamdani motifs images from authentic sources via fieldwork and applied a state-of-the-art method called pix2pix to it. To the best of our knowledge, this dataset is currently the only available dataset of Jamdani motifs in digital format for computer vision research. Our experimental results of the pix2pix model on this dataset show satisfactory outputs of computer-generated images of Jamdani motifs and we believe that our work will open a new avenue for further research.
	\end{abstract}
	
	% ===================
	% # I. Introduction #
	% ===================
	
\section{Introduction}
\begin{figure}[ht] %!t
    \centering
	\begin{subfigure}[b]{0.47\columnwidth}
		\includegraphics[width=1\linewidth]{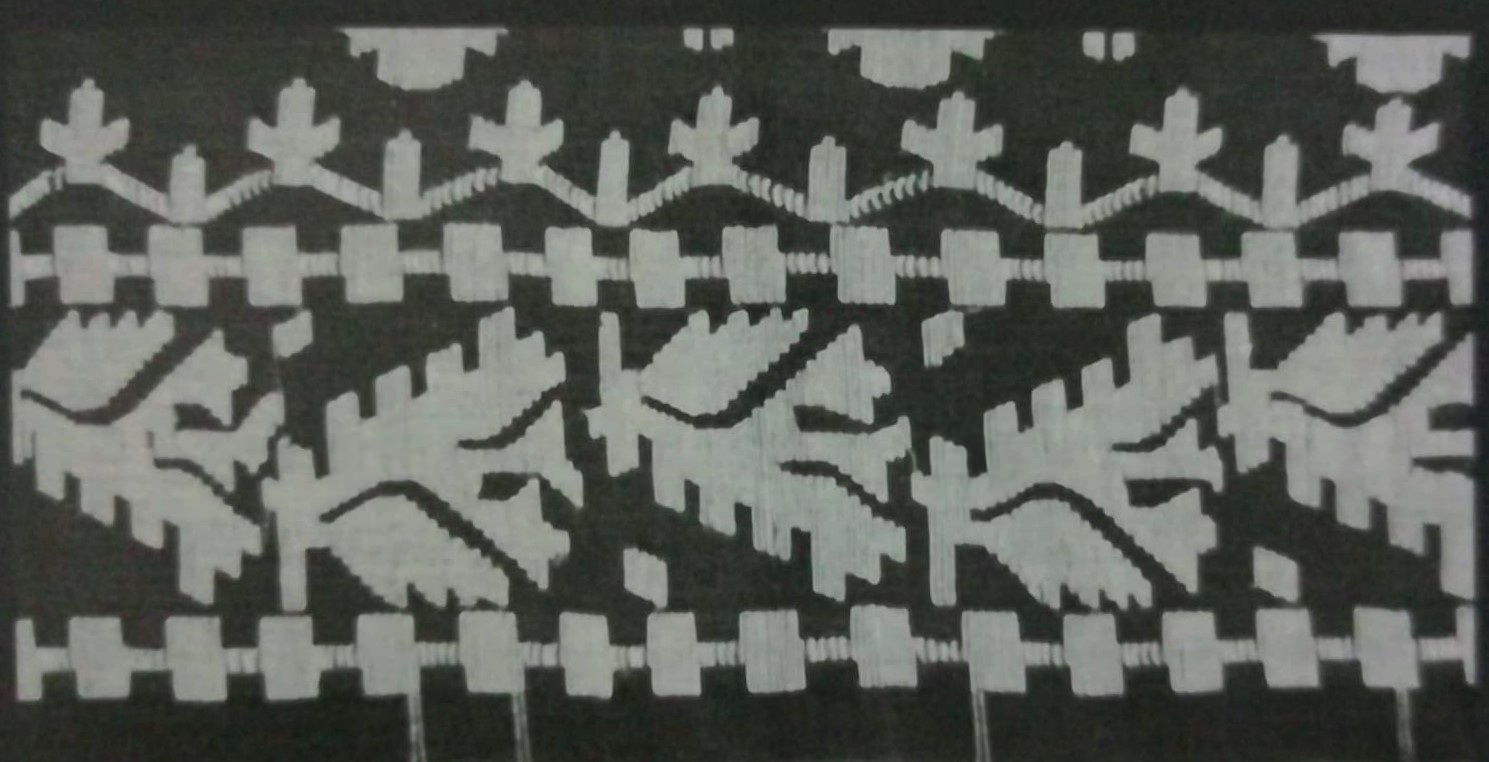}
		\subcaption{Indur Paar}
	\end{subfigure}
	\hfill %%
	\begin{subfigure}[b]{0.47\columnwidth}
		\includegraphics[width=1\linewidth]{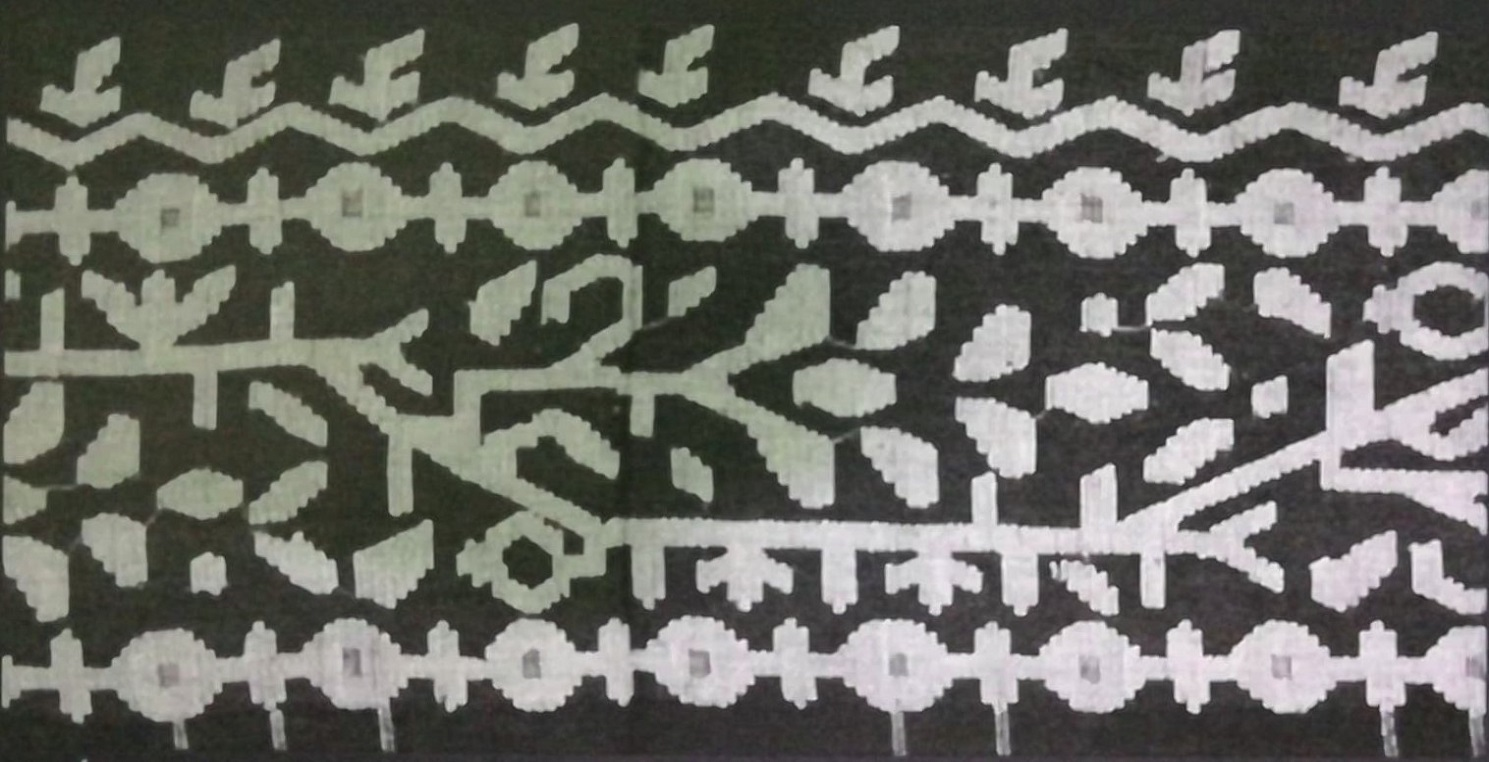}
		\subcaption{Gach Paar}
	\end{subfigure}
	\caption{Various Jamdani Paar. \textit{Source:} \cite{traditional-jamdani}}
	\label{fig:jamdani_par}
\end{figure}

Jamdani is a unique century-old handloom creation embedded in Bangladeshi history having a great socio-economic impact. It is the only surviving fine cotton of the $28$ varieties of Muslin. This handloom piece of cotton---originated from Bengal and Pundra---has a much older history since the third century BC which can be found in the scripture written by the ancient philosopher Chanakya.
%as mentioned by Kautilya having
 \cite{traditional-jamdani}.
In 2013, UNESCO inscribed Bangladesh's Jamdani in the Representative List of Intangible Cultural Heritage of Humanity \cite{jamdani-unesco}.
%Jamdani is an archetypal textile heritage of Bangladesh.
The fact that makes Jamdani an exceptional piece of art is the ornamental motifs; these distinctive opaque forms are consistently woven in geometrical order which creates a poetic play of light and shadow on the sheer pellucid background, with great mastery and craftsmanship. The weaving of Jamdani is an aesthetic process of tedious, labor-intensive, and time-consuming work that is created by artisans of exclusive skills. The geometric patterns are not sketched on the fabric but created from the imagination of the artisans directly on the loom through a mathematical interlacing of warp and weft inspired by the flora and fauna of Bangladesh.  \cite{traditional-jamdani}. 
Figure~\ref{fig:jamdani_par} shows some Jamdani motifs.
%which are sourced from the research work of Mohammad Saidur, a renowned folklorist, BSCIC \& National Craft Council of Bangladesh.

%\begin{figure}[ht] %!t
%	\centering
%	\includegraphics[width=.6\linewidth]{./image/trad_jamdani.jpg}
%	\caption{A traditional Jamdani Saree having different motifs on it}
%	\label{fig:trad_jamdani}
%\end{figure}
%There are certain motifs inspired by the same natural element and have the same name but look different as they were created by different artisans.

\captionsetup[figure]{font=small}

In the past, a Jamdani weaver had to serve both as an artisan and a weaver. The designs were at that time the creation of an artisan's mind which were weaved directly on the loom. There is no alternative to perseverance, concentration, and mathematical calculation while weaving each motif. With time there has been a change in the dedication to work, love for the craft, and urge to continue the traditional profession. For ages, weavers were not properly rewarded for their efforts which made it hard for them to sustain. The discouraging payment and the struggle of artisans with poverty made them shift from this profession while on the other hand, master artisans have passed away. This leads the industry to face a critical phase losing a significant amount of motifs (many complex ones) and designs in the mists of time. Even to these days, each weaver is paid a sum of BDT $14$ per hour for working, and thus they are bound to work $12$ to $14$ hours per day to make an income of around BDT $100$ only which is very demotivating for them. Though in recent days measures are being taken to revive this industry from the verge of extinction, the motifs which got lost in time can never be restored.

In modern times the motifs are no longer confined to textiles only. Nationally and internationally the motifs are being made into splendid materials such as jewelry, home decor, curtains, utensils, etc. On the other hand, there is no Jamdani Artisan alive who weaves the motifs directly on the loom. The weavers now knit the fabric by following designs from catalogs provided by fashion designers. So with the expanding use of motifs, motif creation is no longer confined to the hands of weavers only. Artists, fashion designers, entrepreneurs, and craft enthusiasts are also designing new motifs for creative purposes.
%\begin{itemize}
%	\item It is not financially rewarding for weavers any more. 
%	\item The master weavers are aging. 
%	\item A lot of complex designs are getting lost as weavers these days are not willing to take up such complex task in such low payment. 
%\end{itemize}

\begin{figure}[H]
		\centering
		\includegraphics[width=.8\linewidth, height=2.5cm]{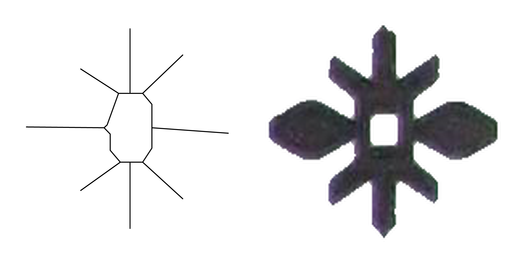}
		\caption{An output of our work: generating Jamdani motif (\textit{right}) from sketch (\textit{left}).}
		\label{fig:intro_img}
	\end{figure}

As such, there are 2 main concerns to be focused on: (1) Resuscitating the industry by motivating the weavers and one way to do it is through launching a collaboration between weavers and designers. (2) The enlarging scope of motifs generation may cause the traditional motifs to get altered. Hence, we take such attempt to develop a system that will mimic the pattern of Jamdani design in the digital images and generate motifs that have the essence of the real ones. This creative machine dedicated to generating Jamdani motifs will bridge the gap between the weavers and designers. This will preserve the motif from further extinction by carrying on the legacy of the surviving motifs. The artificial production of Jamdani motifs will unveil a new source of inspiration for artists by playing the role of an intelligent tool of imagination. Our system is trained in such a way that the visual and artistic appeal of the produced motifs remains intact. In summary, the contributions of our work are listed below.
\begin{itemize}
	\item We generated a dataset called \textbf{Jamdani Noksha} containing images of authentic Jamdani motifs. The dataset is available here: \texttt{\href{https://github.com/raihan-tanvir/generative-jamdani}{https://github.com/ raihan-tanvir/generative-jamdani}}
  \item We presented a motif generation technique that will create a new Jamdani motif from a given stroke.  Our system is trained on our \textbf{Jamdani Noksha} dataset and uses a conditional Generative Adversarial network approach. Our system inputs a rough stroke from the user as an initial guess of the motif (see figure \ref{fig:intro_img}). Being trained, our model can generate entirely new samples of the Jamdani motif which can later be referred for weaving. We investigated different input strokes and demonstrated the outputs.
  
\end{itemize}

	\captionsetup[figure]{font=small}

	% =======================================================
	% # II. Problem Domain #
	% =======================================================
	\section{Background Studies and Related Works}	
	%{Deep Learning}, a sub-field of machine learning which is concerned with algorithms inspired by the construction and activity of the brain called artificial neural networks.
	Firstly, we explain the concept of Generative Adversarial Networks \cite{goodfellow2014generative}---a groundbreaking invention in the field of Deep Learning---since we exploited this in our work.
	The goal of GAN is to synthesize artificial samples, such as images, that are indistinguishable from authentic images.
	%that means it can be used to create anything in any domain like images, music, speech and writing.
	%They are like robotic artists and their outputs are pretty impressive. 
	The basic components of GAN are two neural networks---a generator that synthesizes new samples from scratch, and a discriminator that takes samples from both the training data and the generator’s output and predicts if they are `real' or `fake'.
	The generator input is a random vector or can be stated as noise and therefore its initial output is also noise. Over time, as it receives feedback from the discriminator, it learns to synthesize more realistic images. The discriminator also improves over time by comparing generated samples with real samples, making it harder for the generator to deceive it. 
	Many improvements to the GAN architecture have been proposed through enhancements to the discriminator model with the idea that a better discriminator model will, in turn, lead to the generation of more realistic synthetic images.\\
	
	\textbf{Motivational works:} Below we list different variations of Generative Adversarial Networks which inspired us to use them in our research.
	\begin{itemize}
		\item \textit{pix2pix Generative Adversarial Network}: The \textit{pix2pix} GAN \cite{pix2pix2017} is a general approach for image-to-image translation. It is based on the conditional generative adversarial network, where a target image is generated, conditional on a given input image. \textit{pix2pix} GAN changes the loss function so that the generated image is both plausible in the content of the target domain, and is a plausible translation of the input image.
		
		\item \textit{A Style-Based Generator Architecture for Generative Adversarial Networks}: The Style Generative Adversarial Network, or StyleGAN \cite{stylegan} for short, are an extension to the GAN architecture that proposes large changes to the generator model, including the use of a mapping network to map points in latent space to an intermediate latent space, the use of the intermediate latent space to control style at each point in the generator model, and the introduction to noise as a source of variation at each point in the generator model. The resulting model is capable not only of generating impressively photo-realistic high-quality photos of faces, but also offers control over the style of the generated image at different levels of detail through varying the style vectors and noise.
		
		\item \textit{CycleGAN}: Cross-domain transfer GANs \cite{CycleGAN2017} will be likely the first batch of commercial applications. These GANs transform images from one domain to another domain e.g: real scenery to Monet paintings or Van Gogh painting.
		
		\item \textit{Others:} We got inspired from \cite{sketch-context-comp}, \cite{DBLP:journals/corr/abs-1801-02753}, \cite{salian_2019} as they provide some excellent application of sketch to image translation.
    	We also had our motivation from ESRGAN \cite{Wang_2018_ECCV_Workshops} and PixelDTGAN \cite{Yoo2016PixelLevelDT} as they are some applications of Generative adversarial Networks.
		\end{itemize}	
		\textbf{Similar Work:} In the work titled \textit{`Hand-loom Design Generation using Deep Neural Networks'} \cite{handloomGeneration}, the authors used CycleGAN to generate images aiming to work towards the
		betterment of weavers and industry. Their main target is to translate an image of a normal design to an image of its hand-loomed version. They have created a dataset for this purpose and applied an image-to-image translation to generate new samples.
		Though our approach has similarities with this work in terms of the uniqueness of the dataset, the biggest difference is the domain of both works; as we aim at the Jamdani motif specifically. For this purpose, we have created an entirely different dataset and applied different techniques to generate new motifs.
		Also, our main focus is to create the basic motif from stroke whereas the referenced authors attempted to create a whole design for a better representation of clothes.
		
	% =======================================================
	% # III. Data Collection and Processing #
	% =======================================================
	\section{Our Dataset: \textbf{The Jamdani Noksha}}
	\subsection{Data Collection}
\begin{figure}
		\centering
		\begin{subfigure}[b]{.49\columnwidth}
		    \centering
			\includegraphics[width=1\linewidth , height=3cm]{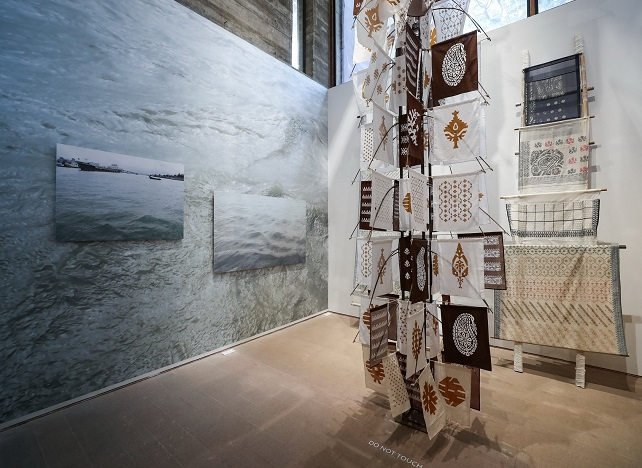}
			%\label{fig:jamdani_festival}
			%\par\bigskip
		\end{subfigure}
		\begin{subfigure}[b]{.49\columnwidth}
		    \centering
			\includegraphics[width=1\linewidth , height=3cm]{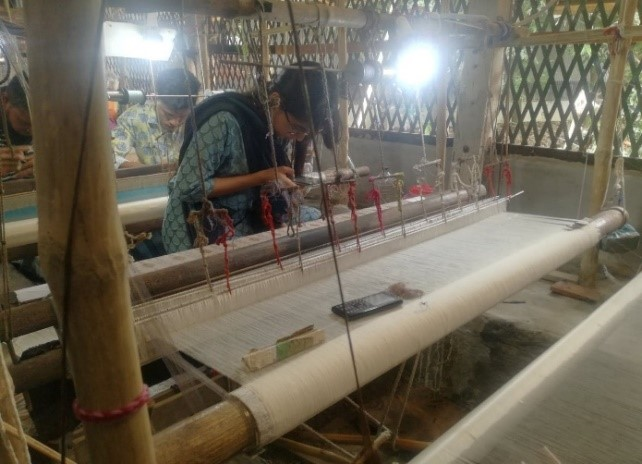}
		%	\label{fig:jamdani_palli}
		\end{subfigure}
		\caption{Data collection from sources such as (\textit{left}) Jamdani Festival 2019 \cite{jamdani-fest}, Bengal Shilpalay, Dhaka , (\textit{right}) Shahina Jamdani Weaving Factory, Jamdani Polli }
		\label{fig:data_source}
	\end{figure}

	As there was no previous data set available for us to work on, the first and foremost step towards building a data set of our own was the visit to Jamdani festival 2019 at Bengal Shilpalay, Dhaka jointly organized by the National Crafts Council of Bangladesh and Bengal Foundation in association with Aarong, Kumudini, Tangail Saree Kutir and Aranya for experiencing the history of Jamdani. The exhibition gave us an overview of the Jamdani industry from the beginning till this day. There we observed the old and new motifs on different sarees and photographs and got introduced to a huge archive enriched with the knowledge of authentic Jamdani designs. We also interviewed the experts working on Jamdani history and culture which enabled us to build basic knowledge about this ancient piece of excellence. The archive is the initial source from where we started collecting our dataset (figure \ref{fig:data_source}).

	Next, we visited \textit{Sonargaon}, the birthplace of Jamdani. Our team was invited to Shahina Jamdani Weaving Factory, one of the oldest and most prominent Jamdani factories of Bangladesh, at Rujganj, Narayanganj, where we experienced the weaving process and unique techniques of the weavers on the loom. From there, we photographed designs weaved on sarees and panjabees for our initial data set. Mr. Juned Ahmad Muhtaseem, an organizing body member of Jamdani Festival 2020, and Mr. Shojib, one of the owners of Shahina Jamdani Weaving Factory are the two contributors who have helped and facilitated us in collecting authentic Jamdani motifs \ref{fig:data_source}.
	
\begin{figure}
		\centering
		\begin{subfigure}[]{.49\columnwidth}
		    \centering
			\includegraphics[width=1\linewidth,height=3cm]{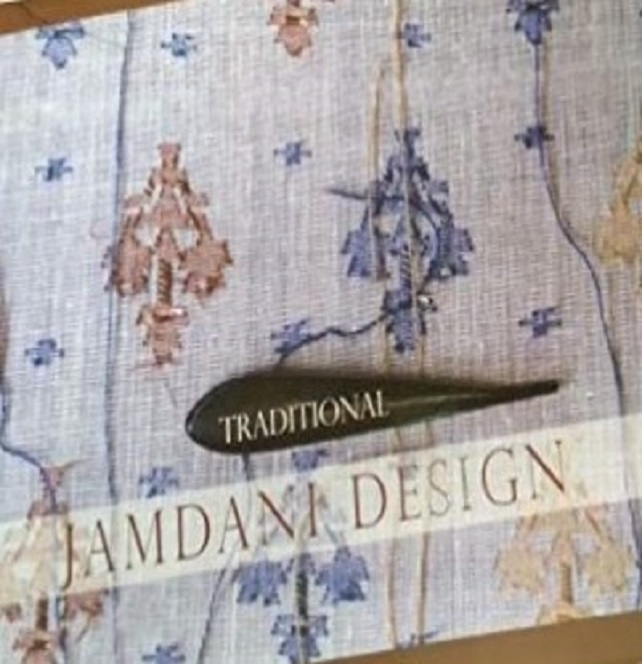}
		\end{subfigure}
		\begin{subfigure}[]{.49\columnwidth}
		    \centering
			\includegraphics[width=1\linewidth,height=3cm]{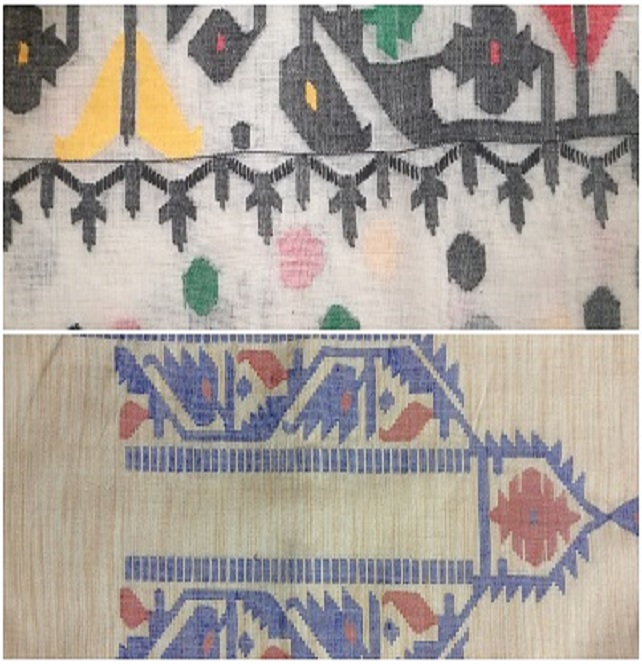}
		\end{subfigure}
		\caption{Data sources book: "Traditional Jamdani Designs” by National Craft Council Bangladesh \cite{traditional-jamdani} (\textit{left}), Sample design collected from a saree and a panjabi while visiting the weaving factory, Rupganj (\textit{right})}
		\label{fig:jamdani_book}
	\end{figure}

\begin{figure}[H] 
		\centering
		\begin{subfigure}[b]{0.49\columnwidth}
		\centering
			\includegraphics[width=1\linewidth, height=2.8 cm]{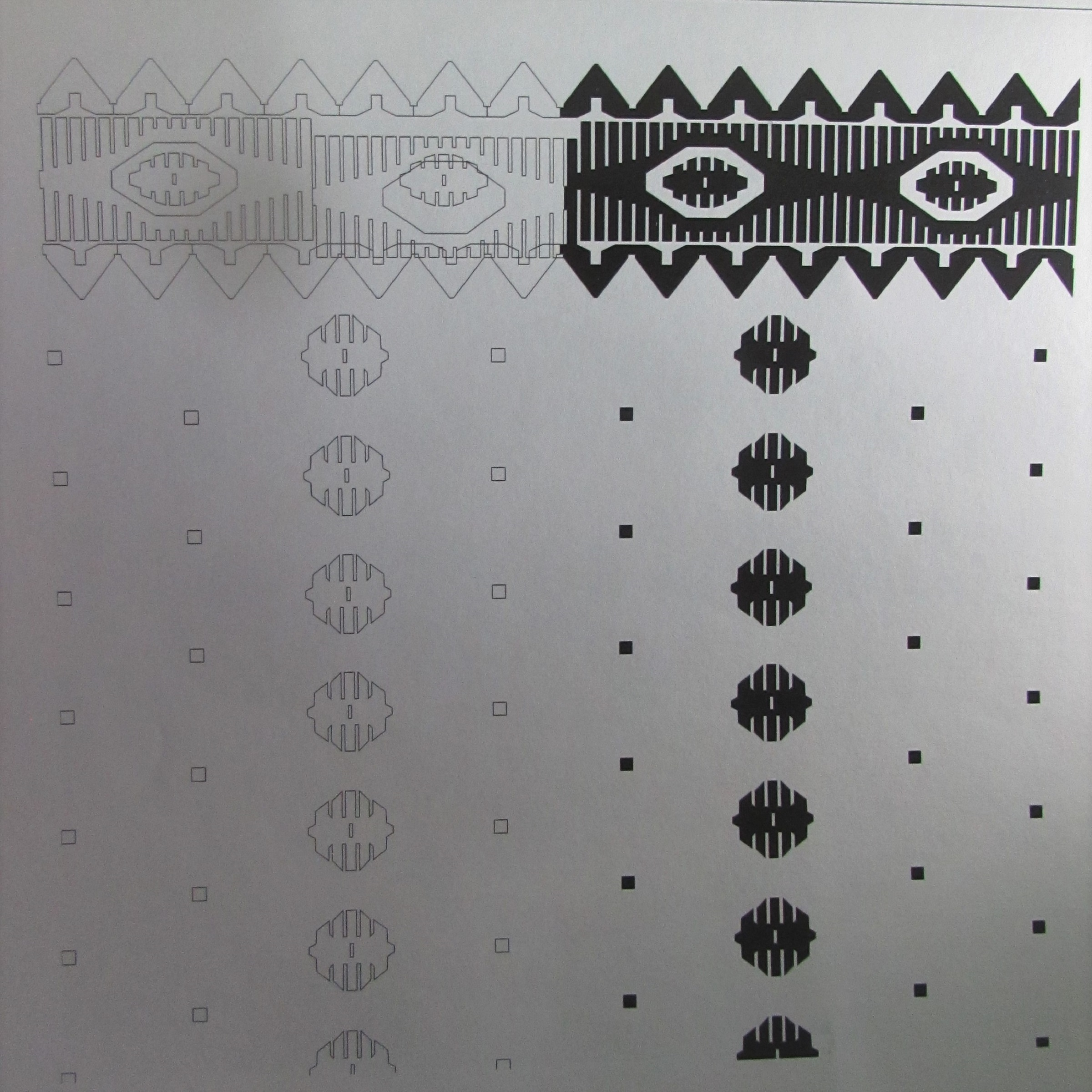}
		\end{subfigure}
		%\hspace{1cm}
		\begin{subfigure}[b]{0.49\columnwidth}
		\centering
		\includegraphics[width=1\linewidth, height=2.8 cm]{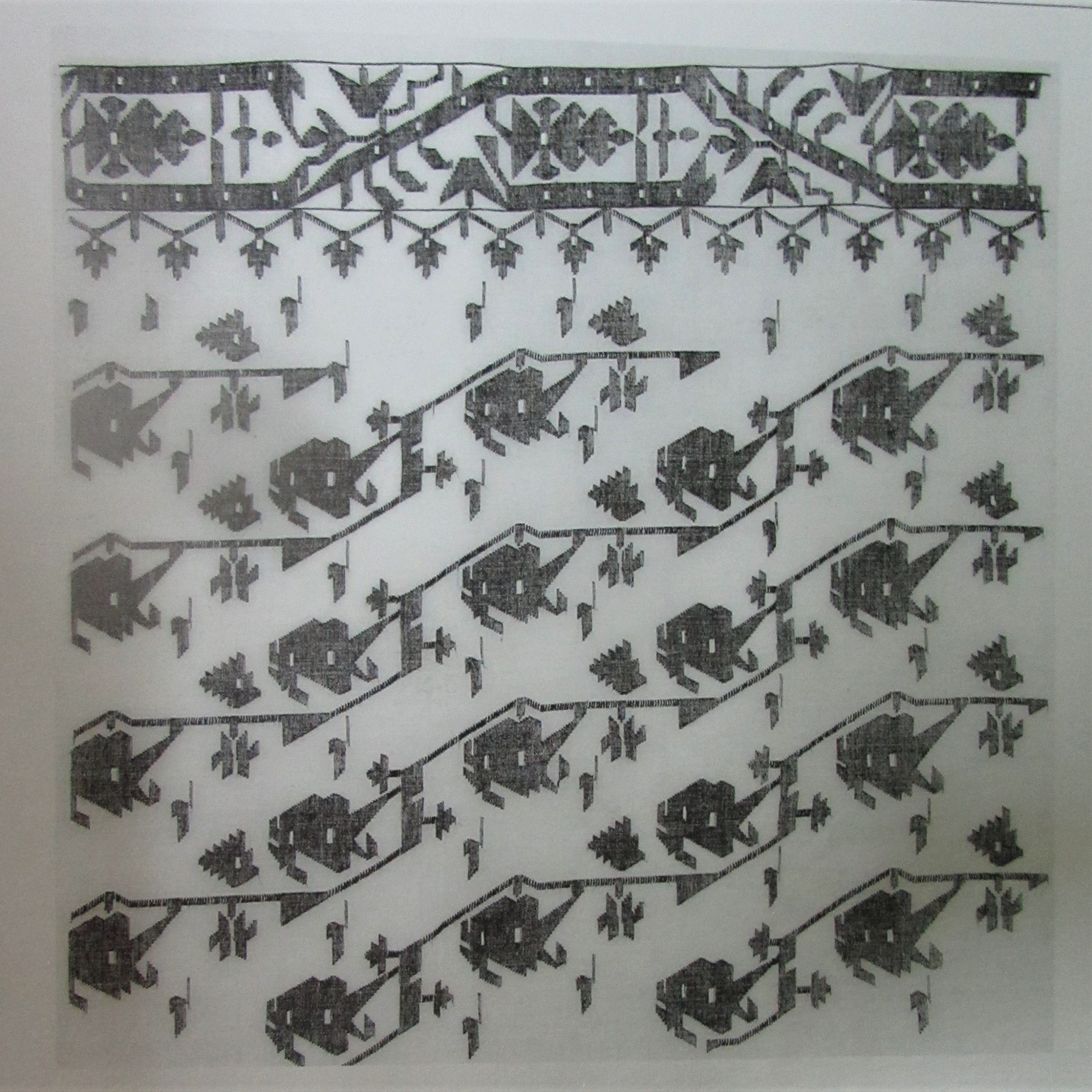}
		\end{subfigure}
		\caption{Some Jamdani designs collected from Traditional Jamdani Book \cite{traditional-jamdani}}
		\label{fig:motif_from_book}
	\end{figure}

	The second source of the raw pictures for \textbf{Jamdani Noksha} dataset was a book called “Traditional Jamdani Designs” by National Craft Council Bangladesh \cite{traditional-jamdani}. It is an archive for authentic traditional Jamdani motifs. This book was published under the project "Preservation of Jamdani Motifs and Designs- an endangered handloom of Bangladesh" supported by Ambassador’s Fund for Cultural Preservation, Embassy of the United States of America. Some sample pictures of different designs are shown in figure~\ref{fig:jamdani_book} and figure~\ref{fig:motif_from_book}.

     %The picture of the book and some

	\captionsetup[figure]{font=small}
	
	\subsection{Data Processing}
	Motifs are the building blocks of Jamdani Designs. Each Jamdani design is a combination of different types of motifs. As we are working with the motifs and adopting \textit{pix2pix}, we had to pre-process the photographs we collected from various authentic sources. Firstly
	every possible Jamdani motif was extracted by cropping the designs into smaller parts.
	%Later, some of the basic image processing techniques were applied to the newly built data set.

	Our target is to generate a motif from a given stroke. Therefore the dataset must include samples as a pair of motifs and strokes so that the model can learn well. However, it is not feasible to include handmade strokes for each of the samples. We found different morphological operations as a replacement for the strokes and we pre-processed our dataset. We built five versions of our dataset---\textit{Skeleton}, \textit{Reduced Branch}, \textit{Sketch}, \textit{Boundary}, and \textit{Enhanced Resolution}. Each sample in the dataset is formatted as a pair of processed and original images according to \textit{pix2pix} \cite{pix2pix2017} %as shown in figure~\ref{fig:data_format}%
	for conditional generative adversarial networks. We explained the processing of this individual version below.
	%Instead of edge, 
	
	\textbf{Skeleton version:} For this version, we consider morphological skeletons as a representation of the motifs.
	%Hence we had to find out the skeletons of the motifs contained in the data set.
	%A dataset of basic skeleton called \textbf{'Skeleton'} was generated by using morphological skeleton operation~\cite{skeletonize1}. 
	%The steps for prodcucing a sample is explained in details below.
	%\paragraph{Real Image to Binary Image}
	%The cropped images collected from the designs are used to generate black and white (binary) images.
	%Before generating binary images
	For this purpose, We applied different morphological image processing like opening, closing, dilation, and erosion on the cropped images according to their needs. Examples of opening and cropping on images are given below in figure \ref{fig:realtobinary2}.

	\captionsetup[figure]{font=small}

	\begin{figure}[H]
	    \centering
		\includegraphics[width=.9\columnwidth]{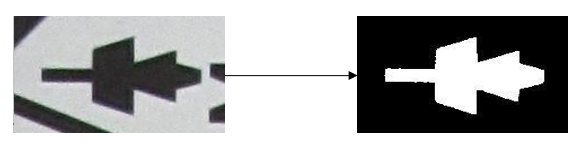}
		\caption{\textit{Left to right:} A jamdani motif image at the left and result of binarizing after applying opening and cropping operations.}
		\label{fig:realtobinary2}
	\end{figure}
	
	%\paragraph{Multiplication on real and binary image}
	%After generating binary image 
	The original image is multiplied with the resulting image obtained from the previous step to remove the background and to keep only the portion of the image containing the motif.
	%The background is also made white to serve the purpose.
%	\paragraph{Generating Skeleton}
	%As skeletons are needed,
	The generated pattern is then used for making a skeleton using the basic skeletonize function \cite{skeletonize1} \cite{skeletonize2}.
	%Skeletonization reduces binary objects to one pixel wide representations.
	It works by making successive passes of the image, where on each pass, border pixels are removed maintaining the constraint that they do not break the connectivity of the corresponding object.
	\captionsetup[figure]{font=small}
		
	\begin{figure}[H]
		\centering
		\includegraphics[width=.9\linewidth]{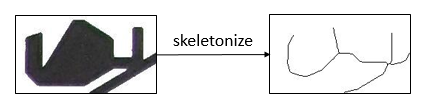}
		\caption{Skeleton (\textit{right}) of a motif in the image}
		\label{fig:skeletonizing}
	\end{figure}
\begin{figure}[H]
		\centering
		\includegraphics[width=1\linewidth]{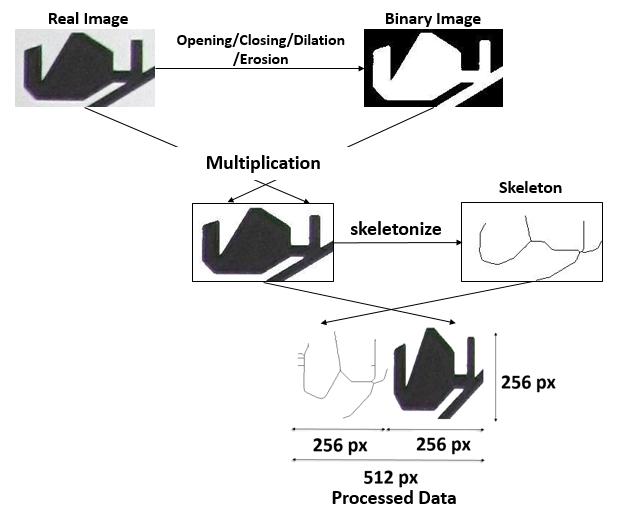}
		\caption{Complete flow of processing for preparing data}
		\label{fig:flow_process}
	\end{figure}
\begin{figure}[H]
		\begin{subfigure}[b]{0.48\columnwidth}
			\frame{\includegraphics[width=\linewidth, height=2cm]{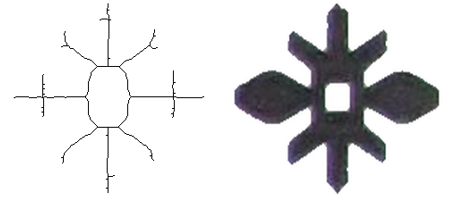}}
		\end{subfigure}
		\hfill %%
		\begin{subfigure}[b]{0.48\columnwidth}
			\frame{\includegraphics[width=\linewidth, height=2cm]{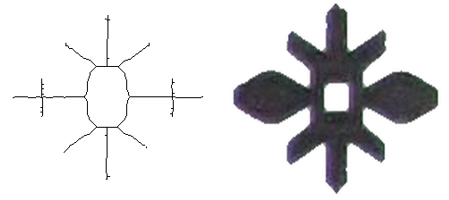}}
		\end{subfigure}
		\caption{Before (\textit{left}) and after (\textit{right}) reduction of the branch. In each pair from left to right, a user-given stroke and the corresponding motif are shown respectively}
		\label{fig:reduced-sketch}
	\end{figure}

	%\paragraph{Merging the skeleton and the motif}
	Finally, the skeleton and the generated pattern are combined side by side and individual data is made. Both the skeleton and generated pattern have the same height ($256$ pixels) and width ($256$ pixel). After the combination, it has a height of $256$ pixels and a width of $512$ pixels.
    %\paragraph*{Flow of processing}
    The complete flow of operation to process data is shown in figure~\ref{fig:flow_process}.
	\captionsetup[figure]{font=small}
	\textbf{Reduced branch version:}
	%After experimenting with the auto generated skeleton we found some kind of problem of extended branching.
	Skeletonization very often provides unwanted branches which mislead the training.
	To solve this, we reduced the extra and unwanted branches using erosion on the binary image and prepared a dataset with reduced branching.
	%called \textbf{'Reduced Branch'}.
	A Difference between the auto-generated and the reduced one is shown in figure~\ref{fig:reduced-sketch}.
	
\begin{figure}[H]
		\begin{subfigure}[b]{0.48\columnwidth}
			\frame{\includegraphics[width=\linewidth]{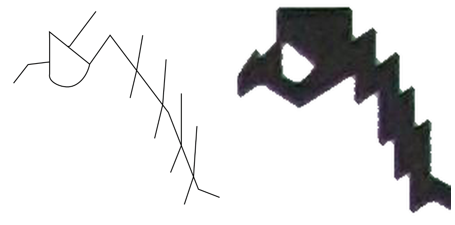}}
		\end{subfigure}
		\hfill %%
		\begin{subfigure}[b]{0.48\columnwidth}
			\frame{\includegraphics[width=\linewidth]{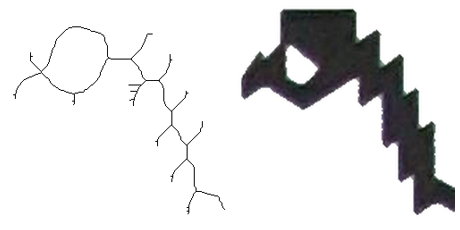}}
		\end{subfigure}
		\caption{Samples of hand-drawn (\textit{left}) and skeleton (\textit{right}) version of our dataset. In each pair from left to right, a user-given stroke and the corresponding motif are shown respectively}
		
		\label{fig:hand-sketch}		
	\end{figure}

	\textbf{Sketch version:} For further experiments to deal with realistic phenomena, we draw the sketches of the skeletons by hand for a more realistic dataset and created a dataset with human annotation. A group of volunteers contributed by sketching the blueprint of the motifs using a digital pen and no morphological operation was performed.
	%called \textbf{'Sketch'}.
	%Here hand drawn skeletons were used to create the dataset.
	 A sample of hand-sketched data with a comparison with the auto-generated one is given in figure~\ref{fig:hand-sketch}.
 	\captionsetup[figure]{font=small}
	 
	\textbf{Boundary version:} We also treated the boundaries of the motifs as input strokes.
	%tried to generate motifs with the help of the boundary of the images. But then we tried the skeleton to make it more realistic. That’s why
	Therefore we created a version of our dataset with the contour of the patterns in images.
	%called \textbf{Boundary}.
	\cite{contourGAN}\cite{sobel}.
	
	\textbf{Enhanced Resolution version:}
	As some photographs were poor in terms of resolution we separated the data with better resolution and prepared another dataset.\\
	%called \textbf{Enhanced Resolution}
	The 5 versions of our dataset and their sizes are given in a table~\ref{tab:dataset-version}.
	
	\begin{table}[H]
		\centering
		\begin{tabular}{@{}ccc@{}}
			\toprule
			SL No. &	Version of Dataset &	Size \\
			\midrule
			1 &	Enhanced Resolution& 1983 \\
			2 &	Reduced Branch  &	913 \\
			3 &	Sketch &	910\\
			4 &	Skeleton & 7932\\
			5 &	Boundary	& 1116 \\
			\bottomrule
		\end{tabular}
		\caption{Sizes of different versions of \textit{Jamdani Noksha} dataset}
		\label{tab:dataset-version}
	\end{table}
	
	% =======================================================
	% # IV. Experiments and Results #
	% =======================================================

	\section{Methodology and Experiments}
\begin{figure}[H]
		\centering
		\begin{subfigure}[b]{.95\columnwidth}
			\includegraphics[width=\linewidth]{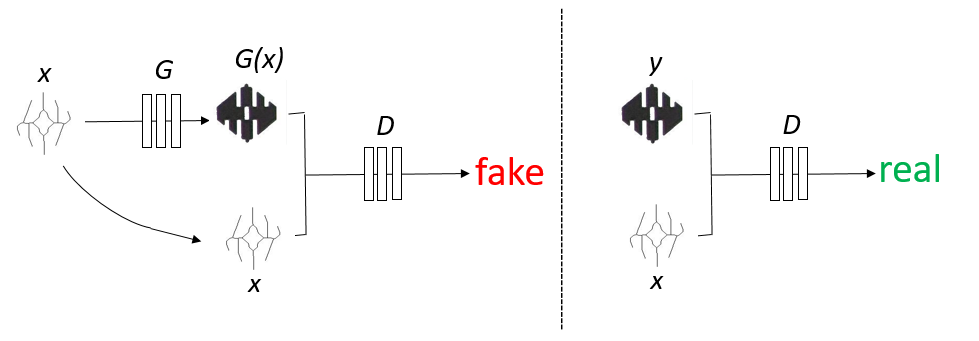}
		\end{subfigure}
		\caption{Training a conditional GAN to map  $skeletons \rightarrow motifs$ }
		\label{fig:gandescription}
	\end{figure}

	We exploited \textit{pix2pix} method for motif generation purposes and train our model accordingly where a generator synthesizes new samples from scratch, and a discriminator takes samples from both the training data and the generator’s output and predicts if the new samples are `real' or `fake'  ~\ref{fig:gandescription}.
	
	\begin{figure*}[ht]
	    \centering
		\begin{subfigure}[b]{0.24\linewidth}
    	    \includegraphics[width=\linewidth]{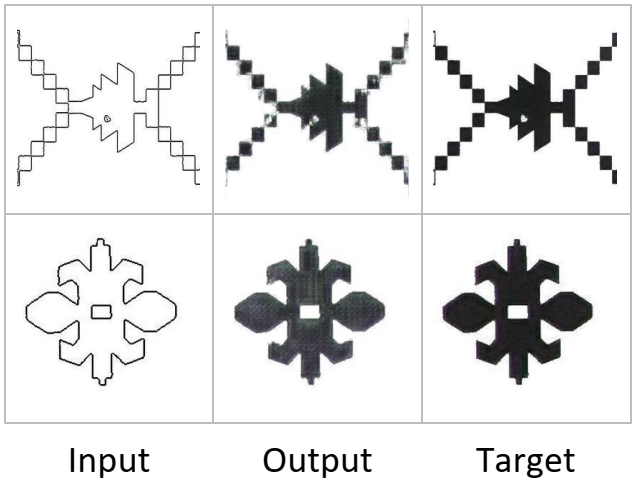}
			\caption{}
			\label{fig:edge2img_sample_o/p}
		\end{subfigure}
		%\hspace{.1cm}
		\begin{subfigure}[b]{0.24\linewidth}
			\includegraphics[width=\linewidth]{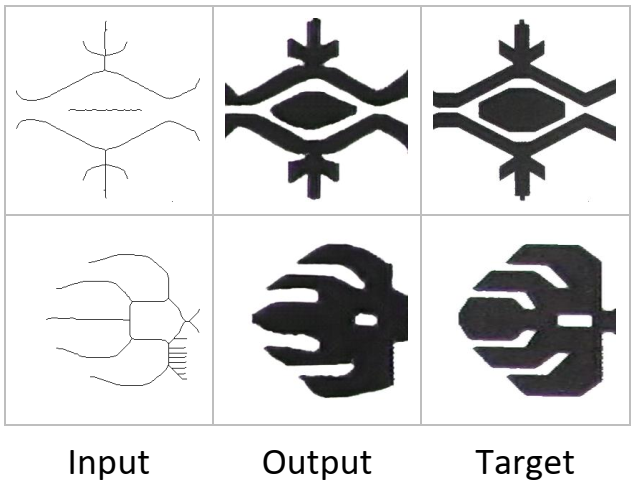}
			\caption{}
			\label{fig:skel2img_sample_o/p}
		\end{subfigure}
		%\hspace{.1cm}
		\begin{subfigure}[b]{0.24\linewidth}
			\includegraphics[width=\linewidth]{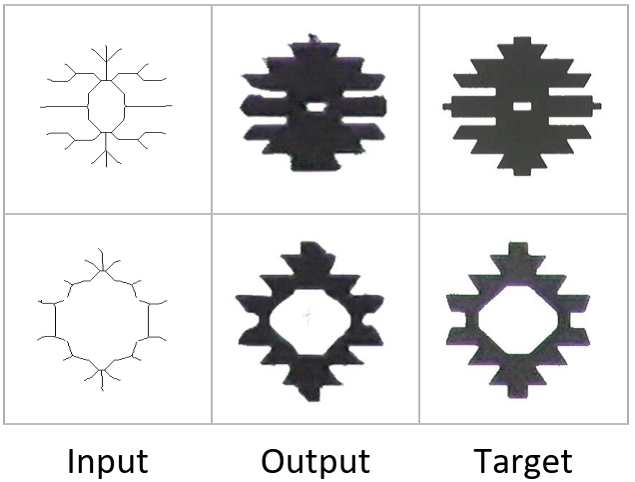}
			\caption{}
			\label{reduce_branch_sample_o/p}
		\end{subfigure}
		%\hspace{.1cm}
		\begin{subfigure}[b]{0.24\linewidth}
    		\includegraphics[width=\linewidth]{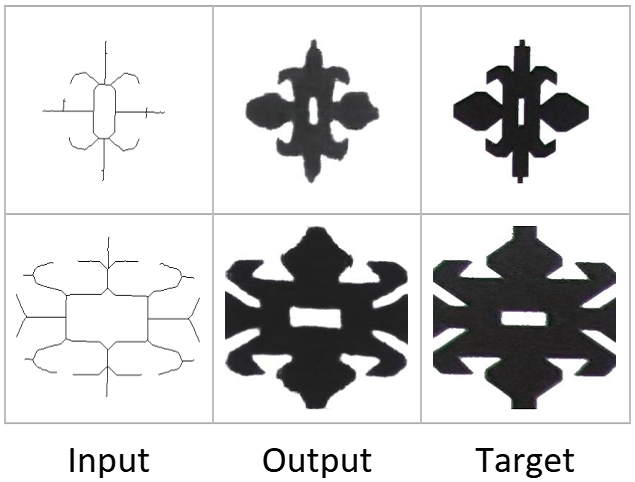}
	    	\caption{}
			\label{fig:all_data_sample_o/p}			
		\end{subfigure}
		
		\caption{Sample output (\textit{middle column of each group}) for a model trained on \textit{Jamdani Noksha}'s (a) Boundary, (b) Enhanced Resolution, (c) Reduced Branch and (d) Skeleton, compared to ground truth (\textit{right column}). \textit{Left column} shows input strokes from the user.}
		\label{fig:sample_o/p}
	\end{figure*}
	The objective of a conditional GAN can be expressed as,
    \begin{equation}
        \begin{aligned}
        \mathcal{L}_{\textit{cGAN}}(G, D)=& \mathbb{E}_{x, \mathbf{y}}[\log D(x,y)]+\\
        & \mathbb{E}_{x, z}[\log (1-D(x, G(x, z))],
        \end{aligned}
    \end{equation}
    where G tries to minimize this objective against an adversarial D that tries to maximize it  ~\cite{pix2pix2017}.  
    The generator is tasked to not only fool the discriminator but
    also to be near the ground truth output in an L2 sense. We
    also explore this option, using L1 distance rather than L2 as
    L1 encourages less blurring  ~\cite{pix2pix2017}:
    \begin{equation}
        \mathcal{L}_{L 1}(G)= \mathbb{E}_{x, y, z}\left[\|y-G(x, z)\|_{1}\right].
    \end{equation}
    Our final objective is
    \begin{equation}
        G^{*}=\arg \min _{G} \max _{D} \mathcal{L}_{\textit{cGAN}}(G, D)+\lambda \mathcal{L}_{L_{1}}(G),
    \end{equation}

	\textbf{Experimental Setup:}
	To explore the generality of \textit{pix2pix} \cite{pix2pix2017}, we followed their two approaches which are boundary $\rightarrow$ image and sketch $\rightarrow$ image.
	Both datasets were divided into training sets, containing $90\%$ of the images, and testing sets, containing $10\%$ of the images. Our technique is implemented in TensorFlow-GPU V1.14.0 and cuDNN V9.0. The experiment has been conducted on a \href{https://colab.research.google.com}{Google Colaboratory}.
	
	\textbf{Experimental Result:}	
	We followed the \textit{pix2pix} \cite{pix2pix2017} for generating the motif using the skeleton. We generated a dataset like the paper having the skeleton and the corresponding image side by side each having $256\times256$ pixels. As the \textit{pix2pix} \cite{pix2pix2017} is a conditional GAN it maps the skeleton as input and the main image as the output.  
	
\begin{figure*}[ht]
	    \centering
		\begin{subfigure}[b]{0.185\linewidth}
    	    \includegraphics[width=\linewidth]{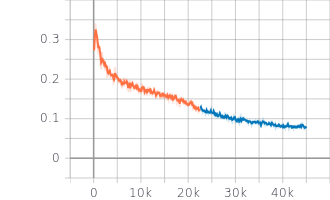}
			\caption{}
			\label{fig:loss_graph_edge2img}
		\end{subfigure}
		\hspace{.1cm}
		\begin{subfigure}[b]{0.16\linewidth}
			\includegraphics[width=\linewidth]{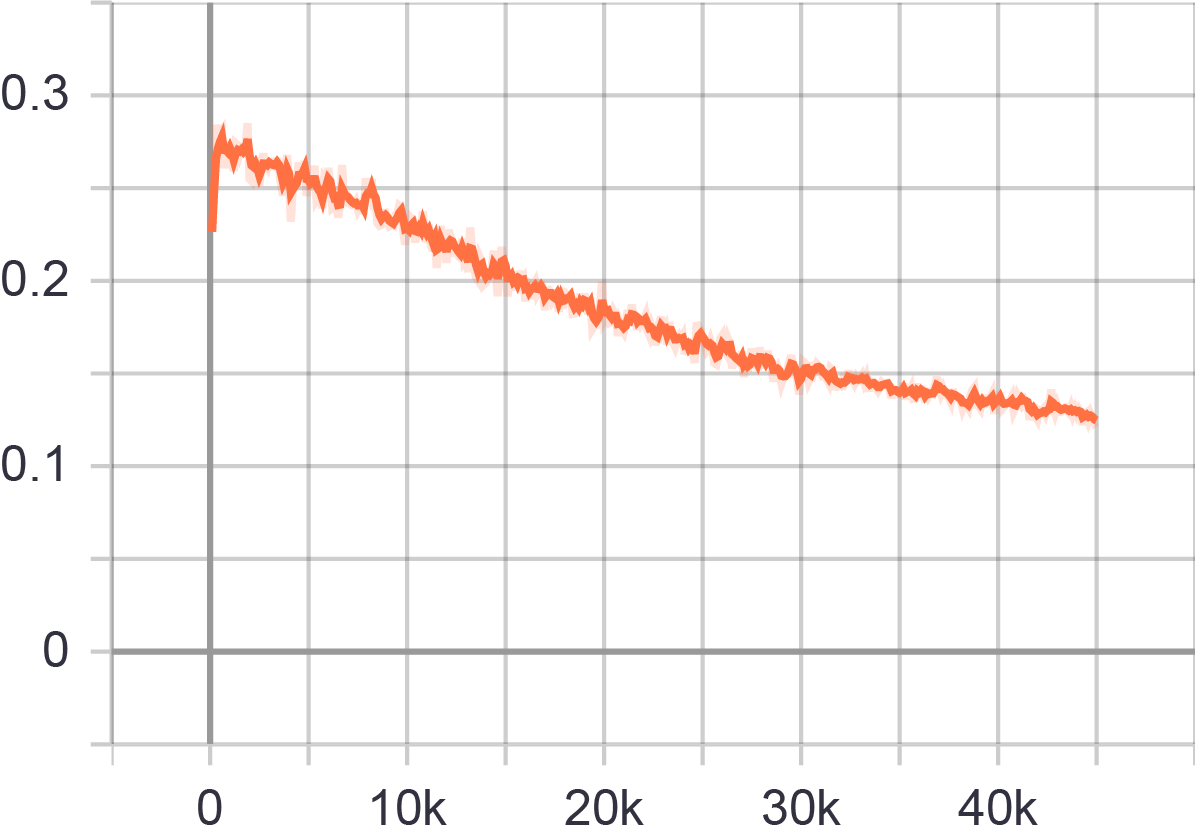}
			\caption{}
			\label{fig:loss_graph_skel2img}
		\end{subfigure}
		\hspace{.1cm}
		\begin{subfigure}[b]{0.17\linewidth}
			\includegraphics[width=\linewidth]{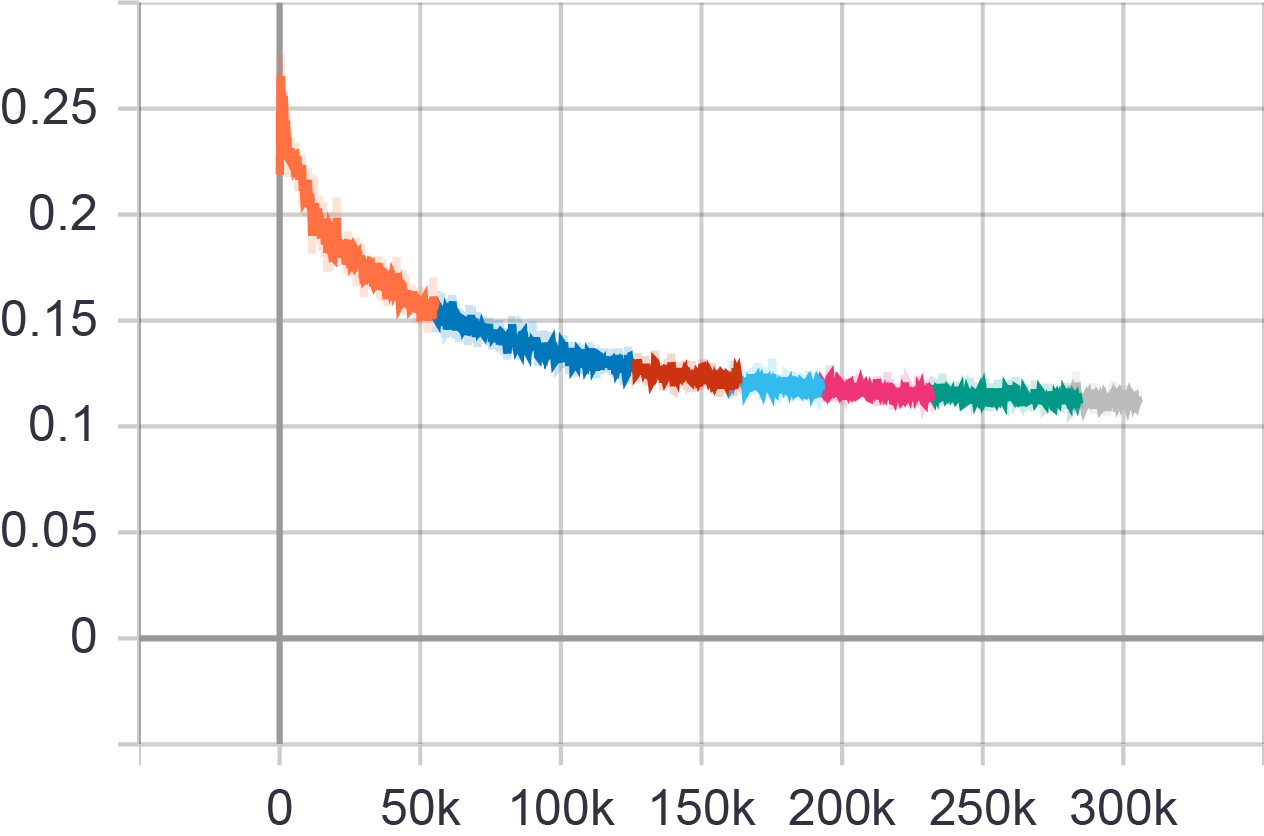}
			\caption{}
			\label{fig:reduce_branch_l1}
		\end{subfigure}
		\hspace{.1cm}
		\begin{subfigure}[b]{0.17\linewidth}
    		\includegraphics[width=\linewidth]{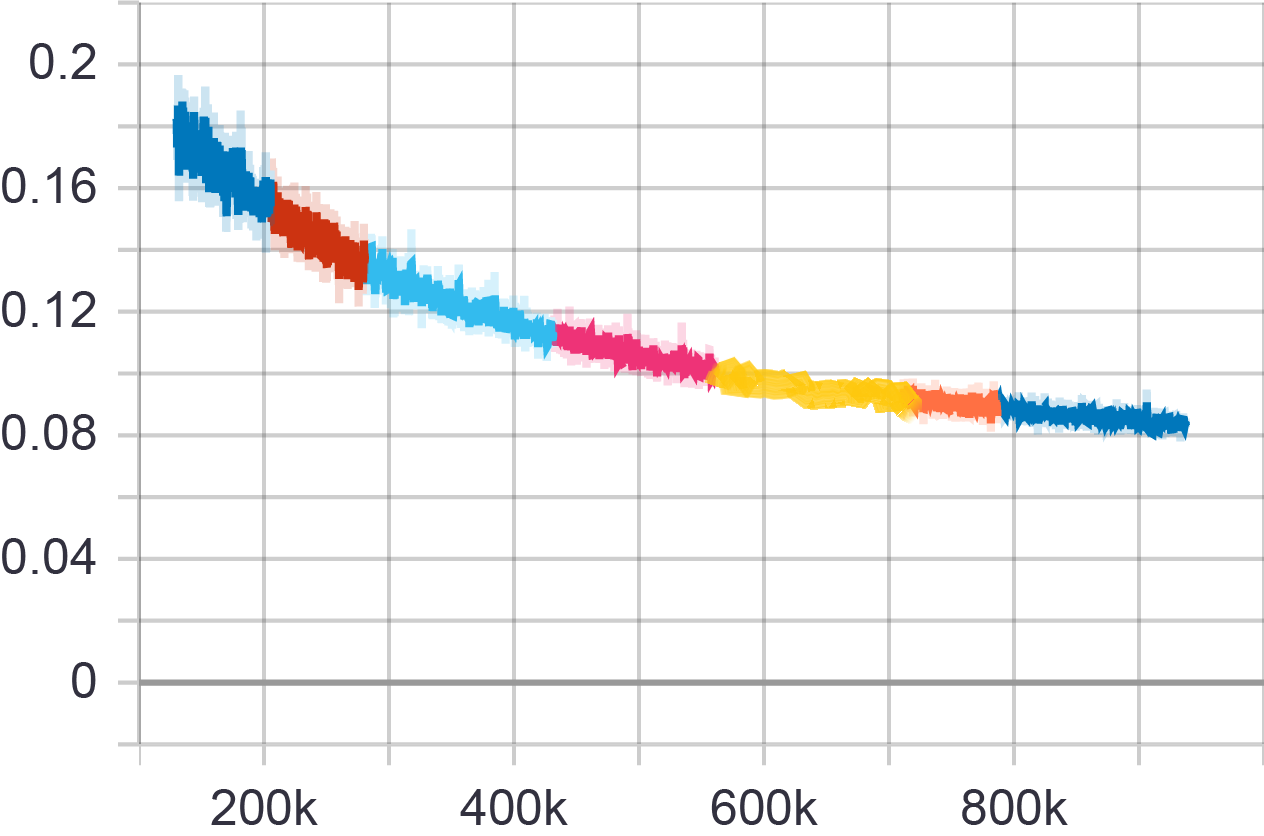}
	    	\caption{}
			\label{fig:all_data_l1}			
		\end{subfigure}
		\hspace{.1cm}
		\begin{subfigure}[b]{0.165\linewidth}
			\includegraphics[width=\linewidth]{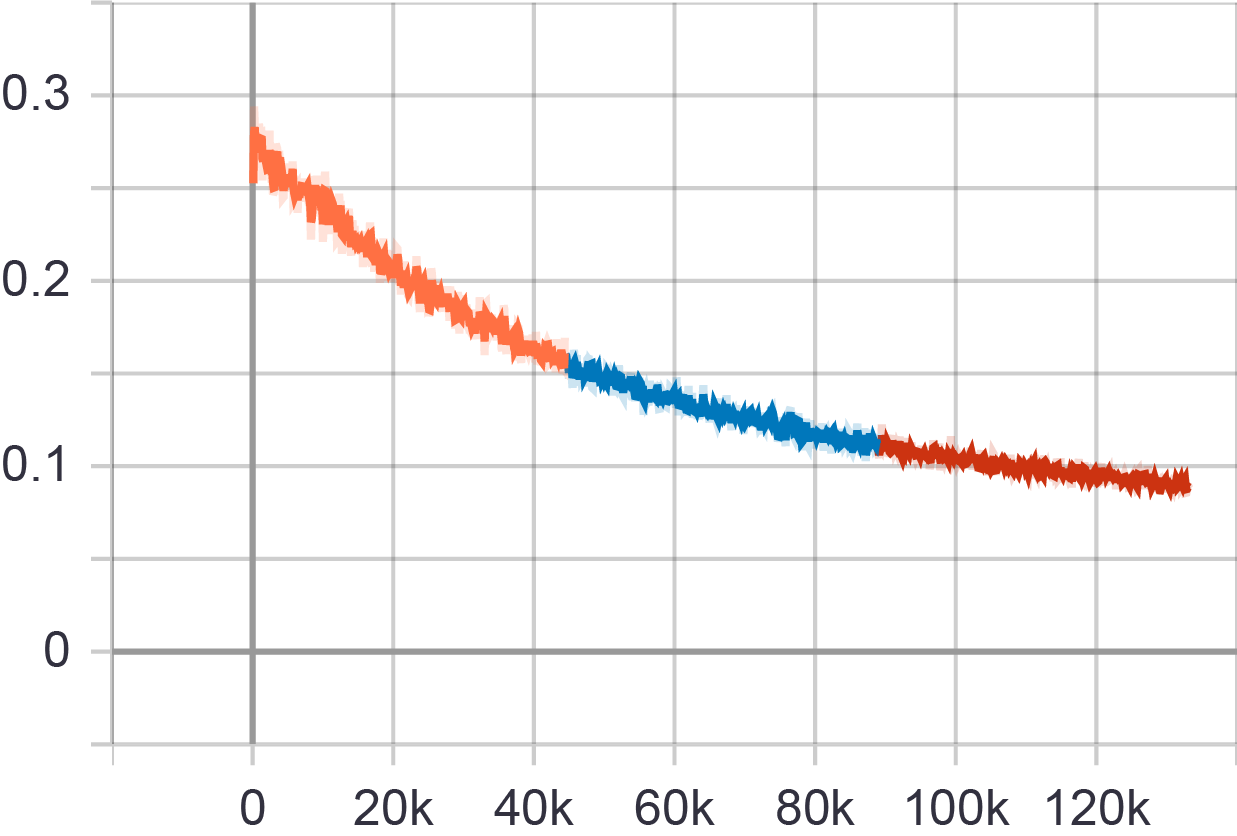}
			\caption{}
			\label{fig:hand_data_l1}			
		\end{subfigure}
		\caption{Loss graph for training on (a) Boundary, (b) Enhanced Resolution, (c) Reduced Branch, (d) Skeleton, and (e) Sketch version}
		\label{fig:graph}
	\end{figure*}

	\textit{Training with \textbf{Boundary} and Output:} Initially, we followed the \textit{Boundary $\rightarrow$ Image} \cite{sobel} approach. We trained the model with the boundary dataset containing $1116$ images with batch size $1$ for $100$ epochs. The training started with a discriminator loss $0.6226$, generator loss $0.908$ and $L1$ loss i.e. mean absolute pixel difference between the target image and generated image $0.2728$, and ended the training with the value $0.8952$, $2.104$ and $0.975$ respectively. See figure~\ref{fig:loss_graph_edge2img}. Some sample outputs are shown in figure~\ref{fig:edge2img_sample_o/p}.
	%graphs
 
\textit{Training with \textbf{Enhanced Resolution} and Output:} After training the model with the \textit{Boundary $\rightarrow$ Image} approach we tried the model with the \textit{Sketch $\rightarrow$ Image} approach. Initially, we prepared the dataset with motifs extracted from the `Traditional Jamdani' book ~\cite{traditional-jamdani}. Our dataset size was $1983$. The training started with a discriminator loss $0.5801$, generator loss $0.9881$, and $L1$ loss $0.2246$ and ended the training with the values $0.8635$, $1.77$, and $0.1239$ respectively. Please see figure~\ref{fig:loss_graph_skel2img}. Sample outputs are shown in figure~\ref{fig:skel2img_sample_o/p}.
	%graphs
 	\captionsetup[figure]{font=small}
	After assessing the generated images by both the approaches it is found that \textit{Sketch $\rightarrow$ Image} produces more plausible images than \textit{Boundary $\rightarrow$ Image} approach. Though loss graphs generated by the tensor board show that \textit{Boundary $\rightarrow$ Image} based training has lower loss metrics than \textit{Sketch $\rightarrow$ Image}, we aim to produce more plausible images concerning the problem domain rather than produce more accurate image-to-image translation i.e. producing images which look like real Jamdani motifs. Taking these facts into account, as well as considering the ease of taking inputs i.e. boundary/sketch of the images from the user for generating the desired Jamdani motif as a key factor, we opted to go for a skeleton-based Jamdani motif generator model with several variations in our dataset. See in table~\ref{tab:dataset-version}.  
	
%	\raggedbottom
	\textit{Training with \textbf{Reduced Branch} and Output:} 
	To make the learning of the model we decided to chop off the branching of the skeletons. The training started with a discriminator loss $0.9196$, generator loss $0.9882$, and $L1$ loss $0.2189$ and ended the training with the values $0.2205$, $5.235$, and $0.11$ respectively. See  figure~\ref{fig:reduce_branch_l1}. Sample outputs are shown in figure~\ref{reduce_branch_sample_o/p}.

	\textit{Training with \textbf{Skeleton} and Output:}
    To gain more plausible output we applied different data augmentation techniques like flip, rotation, etc. to increase the dataset size to $7932$. The training started with a discriminator loss $-1.379$, generator loss $-5.886\times10^{-3}$ and L1 loss $0.1734$ and ended the training with the values $-1.385$, $-1.5\times10^{-4}$ and $0.08392$ respectively. See  figure~\ref{fig:all_data_l1}. Sample outputs are shown in figure~\ref{fig:all_data_sample_o/p}.
	%all data o/p
 	
 	\captionsetup[figure]{font=small}
	\begin{figure}
		\centering
		\includegraphics[width=.7\linewidth]{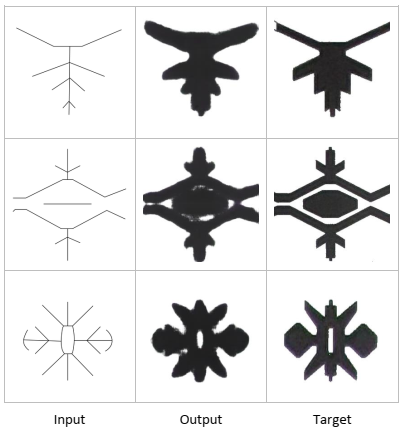}
		\caption{Sample output (\textit{middle column}) for model trained on \textit{Jamdani Noksha}'s hand-drawn sketch version, compared to ground truth (\textit{right column}). \textit{Left column} shows input strokes from the user.}
		\label{fig:hand_data_sample_o/p}
	\end{figure}

	\textit{Training with \textbf{Sketch} and Output:} 
	Finally, we chose to draw a handcrafted sketches so that the model gets to learn the actual input sketch to output mapping. We created $250$ sketches corresponding to our target output with the help of Adobe Illustrator and increase the dataset size four times by applying data augmentation. The training started with a discriminator loss $-0.7185$, generator loss $-0.1069$, and $L1$ loss $0.2526$ and ended the training with the values $-1.362$, $-0.0173$, and $0.0885$ respectively. See figure~\ref{fig:hand_data_l1}. Sample outputs are shown in figure~\ref{fig:hand_data_sample_o/p}.
	%handsketch o/p

	\section{Limitations And Future Work}
	We move forward with a view that our unique dataset which represents our culture and the cutting-edge technology of GAN will intersect and bring out a new dimension in both the areas of research and tradition. We hope our contribution adds to the industry and economy at the greater end. To our knowledge, the research work that we have initiated is a unique and one-of-a-kind contribution to this field and it holds the potential to go further. But the opportunities are not without limitation. The biggest constraint is the insufficiency of data. To have a more structured and efficient model more data is needed. Below are the hurdles to overcome and the scopes to explore is discussed:

	\begin{itemize}
		\item	Make the outputs more realistic and flawless.
		\item	Classify the Jamdani motifs/patterns from other designs or patterns.
		\item	Generate larger designs using the building block motifs. 
		\item	Convert different objects into a geometric pattern that resembles the hand-loomed Jamdani designs.
	\end{itemize}

	\section{Conclusion}
	To contribute to the continuation of the Jamdani tradition of Bangladesh we believe the most reliable solution is to digitize this industry in an artificially intelligent manner.
	In this paper, we presented an approach to generate Jamdani motifs based on initial guesses from users' strokes.
	This will not only ensure the preservation of the most complicated and oldest motifs but will also help the artisan involved in sector work more efficiently and easily. 
	We believe our research will serve as a foundation for more dynamic and creative work in this domain that will assure the development of the Jamdani industry by helping weavers sustain in an enriched economic condition. The core identity of the Jamdani tradition has to be kept unchanged. Our \textbf{Jamdani Noksha} Dataset is not only a collection of motifs of different eras but a rich source for the researchers enabling them to unfold the unknown.  
	
	% ==================
	% # ACKNOLEDGMENTS #
	% ==================
	
	%\section*{Acknowledgment}
	
	% ==============
	% # REFERENCES #
	% ==============
	\bibliographystyle{unsrt}
	\bibliography{ref}

\begin{thebibliography}{10}

\bibitem{traditional-jamdani}
Chandra~Shekhar Shaha.
\newblock {\em Traditional Jamdani Design}.
\newblock Bangladesh National Museum, March 1, 2018.

\bibitem{jamdani-unesco}
Traditional art of jamdani weaving.
\newblock
  \url{https://ich.unesco.org/en/RL/traditional-art-of-jamdani-weaving-008 },
  2013 (accessed September 5, 2020).

\bibitem{goodfellow2014generative}
Ian~J. Goodfellow, Jean Pouget-Abadie, Mehdi Mirza, Bing Xu, David
  Warde-Farley, Sherjil Ozair, Aaron Courville, and Yoshua Bengio.
\newblock Generative adversarial networks, 2014.

\bibitem{pix2pix2017}
Phillip Isola, Jun-Yan Zhu, Tinghui Zhou, and Alexei~A Efros.
\newblock Image-to-image translation with conditional adversarial networks.
\newblock {\em CVPR}, 2017.

\bibitem{stylegan}
Tero Karras, Samuli Laine, and Timo Aila.
\newblock A style-based generator architecture for generative adversarial
  networks.
\newblock {\em CoRR}, abs/1812.04948, 2018.

\bibitem{CycleGAN2017}
Jun-Yan Zhu, Taesung Park, Phillip Isola, and Alexei~A Efros.
\newblock Unpaired image-to-image translation using cycle-consistent
  adversarial networks.
\newblock In {\em Computer Vision (ICCV), 2017 IEEE International Conference
  on}, 2017.

\bibitem{sketch-context-comp}
Yongyi Lu, Shangzhe Wu, Yu-Wing Tai, and Chi-Keung Tang.
\newblock Sketch-to-image generation using deep contextual completion, 11 2017.

\bibitem{DBLP:journals/corr/abs-1801-02753}
Wengling Chen and James Hays.
\newblock Sketchygan: Towards diverse and realistic sketch to image synthesis.
\newblock {\em CoRR}, abs/1801.02753, 2018.

\bibitem{salian_2019}
Isha Salian.
\newblock Gaugan turns doodles into stunning, realistic landscapes: Nvidia
  blog.
\newblock
  \url{https://blogs.nvidia.com/blog/2019/03/18/gaugan-photorealistic-landscapes-nvidia-research},
  Oct 2019.

\bibitem{Wang_2018_ECCV_Workshops}
Xintao Wang, Ke~Yu, Shixiang Wu, Jinjin Gu, Yihao Liu, Chao Dong, Yu~Qiao, and
  Chen Change~Loy.
\newblock Esrgan: Enhanced super-resolution generative adversarial networks.
\newblock In {\em Proceedings of the European Conference on Computer Vision
  (ECCV) Workshops}, September 2018.

\bibitem{Yoo2016PixelLevelDT}
Donggeun Yoo, Namil Kim, Sunggyun Park, Anthony~S. Paek, and In-So Kweon.
\newblock Pixel-level domain transfer.
\newblock In {\em ECCV}, 2016.

\bibitem{handloomGeneration}
R.~K {Bhattacharjee}, M.~{Nandi}, and A.. {Jha}.
\newblock Handloom design generation using deep neural networks.
\newblock
  \url{https://github.com/rajatkb/Handloom-Design-Generation-using-Deep-Neural-Networks},
  2018.

\bibitem{jamdani-fest}
{\em Jamdani Festival}, 2019 (accessed August 23, 2020).
\newblock \url{https://jamdanifestival.com}.

\bibitem{skeletonize1}
Liping Yang, Diane Oyen, and Brendt Wohlberg.
\newblock A novel algorithm for skeleton extraction from images using
  topological graph analysis, 06 2019.

\bibitem{skeletonize2}
J.~Ma, Tsviatkou Yurevich, and V.~Kanapelka.
\newblock Image skeletonization based on combination of one- and
  two-sub-iterations models.
\newblock {\em Informatics}, 17:25--35, 06 2020.

\bibitem{contourGAN}
Hongju Yang, Yao Li, Xuefeng Yan, and Fuyuan Cao.
\newblock Contourgan: Image contour detection with generative adversarial
  network.
\newblock {\em Knowledge-Based Systems}, 164, 10 2018.

\bibitem{sobel}
Wenshuo Gao, Xiaoguang Zhang, Lei Yang, and Huizhong Liu.
\newblock An improved sobel edge detection.
\newblock {\em Proceedings - 2010 3rd IEEE International Conference on Computer
  Science and Information Technology, ICCSIT 2010}, 5:67 -- 71, 08 2010.

\end{thebibliography}

\end{document}